\documentclass[12pt,twocolumns]{IEEEtran}
\pdfoutput=1

\usepackage{amsmath}
\usepackage{array}
\usepackage{cite}
\usepackage{amsfonts}
\usepackage{amssymb}
\usepackage{graphicx} % for pdf, bitmapped graphics files
\usepackage{epstopdf}
\usepackage{float}
\usepackage{algpseudocode}
\usepackage{balance}

\usepackage[fancythm,fancybb]{jphmacros2e} 
\usepackage{amsfonts}
\usepackage{mathrsfs}
\usepackage{amsmath}
\usepackage{amsthm}
\usepackage{array}
\usepackage{comment}
\usepackage{tabularx}

 \usepackage[norefs,nocites]{refcheck}

\begin{document}

\algdef{SE}[FOR]{NoDoFor}{EndFor}[1]{\algorithmicfor\ #1}{\algorithmicend\ \algorithmicfor}
\algdef{SE}[IF]{NoThenIf}{EndIf}[1]{\algorithmicif\ #1}{\algorithmicend\ \algorithmicif}%
\algdef{SE}[WHILE]{NoDoWhile}{EndWhile}[1]{\algorithmicwhile\ #1}{\algorithmicend\ \algorithmicwhile}%
\def\NoNumber#1{{\def\alglinenumber##1{}\State #1}\addtocounter{ALG@line}{-1}}

\IEEEoverridecommandlockouts
	\allowdisplaybreaks
	
\title{A Multi-step and Resilient Predictive Q-learning Algorithm for IoT with Human Operators in the Loop: A Case Study in Water Supply Networks}

\author{Maria Grammatopoulou,
        %John~Doe,~\IEEEmembership{Fellow,~OSA,}
        ~Aris Kanellopoulos, % <-this % stops a space
        ~Kyriakos G.~Vamvoudakis \\ 
        and
        ~Nathan Lau
% <-this % stops a space
\thanks{M.~Grammatopoulou and N.~Lau are with the Grado Department of Industrial and Systems Engineering, Virginia Tech, Blacksburg, VA, $24061$, USA e-mail: (mariagr@vt.edu, nathan.lau@vt.edu).}
\thanks{A.~Kanellopoulos and K.~G.~Vamvoudakis are with the Daniel Guggenheim School of Aerospace Engineering, Georgia Institute of Technology, Atlanta, GA, $30332$, USA e-mail: (ariskan@gatech.edu, kyriakos@gatech.edu).}}

\maketitle
 \thispagestyle{empty}
\pagestyle{empty}

\begin{abstract}
We consider the problem of recommending resilient and predictive actions for an IoT network in the presence of faulty components, considering the presence of human operators manipulating the information of the environment the agent sees for containment purposes. 
The IoT network is formulated as a directed graph with a known topology whose objective is to maintain a constant and resilient flow between a source and a destination node. 
The optimal route through this network is evaluated via a predictive and resilient Q-learning algorithm which takes into account historical data about irregular operation, due to faults, as well as the feedback from the human operators that are considered to have extra information about the status of the network concerning locations likely to be targeted by attacks. 
To showcase our method, we utilize anonymized data from Arlington County, Virginia, to compute predictive and resilient scheduling policies for a smart water supply system, while avoiding (i) all the locations indicated to be attacked according to human operators (ii) as many as possible neighborhoods detected to have leaks or other faults. This method incorporates both the adaptability of the human and the computation capability of the machine to achieve optimal implementation containment and recovery actions in water distribution.
\end{abstract}


\section{Introduction}
According to the World Health Organization, urban residents account for $54\%$ of the total global population. 
This means that cities face challenges in meeting demands of resources such as energy and water, whose availability depends on factors including climate change, weather patterns and natural and man-made hazards. 
Many cities around the world are already overcrowded leading to traffic congestion, and a strain on resources such as water, energy and safe housing.
Municipal governments around the globe are beginning to recognize that big data and Internet of Things (IoT) can contribute to developing sustainable connected communities while improving the daily life of their citizens. 
%Meeting sustainability goals of cities requires completely new concepts for urban mobility and sustainable use of resources, evolving a typical city into a smart one. 
%The smart city concept is based on ubiquitous sensing and connected technology that communicating ``live'' information amongst connected ``things''.

Given the exponential increase of devices equipped with networking capabilities, researchers have introduced the concept of the IoT \cite{atzori2010internet}. 
The IoT consists of both physical and cyber devices communicating via standard TCP/IP protocols. 
%IoT or connected devices are projected to reach $24$ billion by $2020$ \cite{bassi2008internet}, and are extremely heterogeneous in their applications. 
%As an example, we can consider a single network containing smart meters deployed in public areas \cite{zanella2014internet}, individual wearable devices collecting health data \cite{gubbi2013internet}, as well as large-scale systems like heating, ventilation, and air conditioning (HVAC)\cite{alfuqaha2015internet}, all exchanging data with a single user's mobile phone. 

%The IoT protocols not only will have to support the different big data structures, but can also have a direct effect on the physical world and thus public safety in many cases. 
%Consequently, it is important to ensure safe and robust operation of the IoT networks in case of random or malicious faults.

One area of the IoT applications in smart cities is water distribution, which must be operational at all times, even in the presence of faults and, thus, continuous monitoring of the network status is a priority. 
However, human analysts can process only a limited amount of numerical data at a time although they are very adaptive to novel situations or unanticipated events. 
Further, human operators and analysts are often preoccupied with other duties (e.g., supporting maintenance).
Machine learning (ML) methods can be valuable to augment human monitoring given their real time computational abilities, and can help make some auxiliary decisions for human operators \cite{liang2017internet}.

The introduction of ML and tools of cognitive engineering into traditional systems can facilitate and speed up the decision making and performance of the systems. 
 Cognitive engineering is an interdisciplinary field concerned with analysis, design, and evaluation of complex systems of people and technology, and combines knowledge from human factors, systems engineering, cognitive engineering, and human-computer interaction design \cite{gersh2005internet}. 
Effective human-system integration in systems engineering help capitalize on human potential, while circumventing human limitations. 

Water systems are designed to operate both consistently and economically, and to deliver water in sufficient quantity, of acceptable quality, and at appropriate pressure \cite{jung2007internet}. 
%Industrial control systems adopting the Supervisory Control and Data Acquisition (SCADA) architecture play a key role in operating water distribution networks. 
The data collected by the sensors in the network are transmitted to programmable logic controllers regulating the equipment settings, and to other client computers displaying process information to the human operators. 
These data are presented to the human operators to decide whether manual control is necessary, overriding the automatic systems that operate normally. 
%SCADA systems play an important role in operating the equipment and controlling the process automatically, as well as presenting process information for human to understand and intervene the process accordingly. 

The focus of our work is to examine how ML algorithms in Supervisory Control and Data Acquisition (SCADA) systems can be engineered to incorporate human decisions and actions in optimizing policy in the long run. 
This particular study examines reinforcement learning for determining the water flow in the network when preventative measures need to be taken by the human due to unanticipated events. 
Unanticipated events not only include cyber-attacks but also unknown environmental factors (e.g., obscure contamination) or physical/plant failures (e.g., sensor failures) in the water systems. 
Both intentional and unintentional contaminant threats are top vulnerabilities of the physical attributes of the water systems that can pose significant safety ramifications. 

\subsection{Structure}
The paper is organized as follows. 
The next section presents a general description of the predictive and resilient Q-learning algorithm, and the two methods for having humans in the loop to be used. 
In the third section, we apply the proposed algorithms to a water supply network utilizing real data of observed leaks and simulating the stochastic changes of the network status that the human operator observes. 
Finally, the fourth section discusses our findings and future research directions.

\section{Problem Formulation and Proposed Framework}

\subsection{Network Model}
We model the IoT subnetwork of the water distribution system as a directed graph, which allows us to describe the data transfer between physical networks monitored and controlled through the IoT, such as a smart city pipeline network. 
Expert analysts are expected to supervise the smooth and safe operation of different sub-networks in future IoT and smart cities.
Thus, algorithms must be developed to facilitate joint human-machine decision making, by processing the raw data collected and suggesting control actions to the human operators, who can then interact with and influence the course of automatic controllers. 
Hence,the human operators maintain high-speed situational awareness of the network status. 
One approach is reinforcement learning that can predict future behavior based on previous data, and extract recommended policies, attack trends and vulnerability assessments while allowing the human to be in the loop and intervene as necessary.

\subsection{Predictive Q-learning}
We utilize a Q-learning algorithm to derive the optimal policies that will be recommended as the resilient policies for the network operator. 
Similar to other Q-learning approaches, we define an action-dependent value function $Q^k(s,a)$ for state $s$ and action $a$.
This function should also contain information about the measured past faults. 
Therefore, the Q-function has the form,
\begin{equation}\label{eq:q}
Q^k(s,a) = \sum_{t=k-M}^{k}R^t +  \sum_{t=1}^{k-M} b^{t} R^t,
\end{equation}
where $R^k$ is the cost to transfer a single packet from node $i$ to node $j$ at time $k$, and is defined as $R^k = [r_{ij}^k] \textrm{, }r_{ij}^k\geq 0 \textrm{ }\forall i,j,k$, and $b\in(0,1)$ is the discount factor and $M>0$ is the size of the window. 
The cost matrix, $R^k$, is dynamically updated to reflect data collected during the run of the system, thus is containing different values at each time $k$.
%Equation \eqref{eq:q} implies that, during each time-step, we take into account the faults observed throughout a predefined time window of length $M$, while knowledge observed farther in the past affects the scheduling less.
Finally, our objective is to derive the optimal Q-function,
\begin{equation*}
Q_\textrm{opt}^k = \min_a Q^k(s,a).
\end{equation*}

The dynamic nature of the environment, caused by faults and attacks on the network, leads to a shifting structure of the reward matrix and the Q-function as described above. 
It is important for the decision maker to have the ability to adapt the scheduling policy as fast as possible. 
For this reason, the Q-learning problem in this work is inspired by the structure of Predictive Q-routing \cite{choi1996internet}. 
The novelty of PQ-routing lies in the fact that it accounts for the congestion created by the optimal routing policy itself. 
Specifically, it is argued that when we statically use the optimal path, the increase in traffic from specific nodes decreases the efficiency.

\begin{comment}
While simple Q-learning utilizes the principles of reinforcement learning by exploring the state and action spaces and gradually converging to the optimal policy, predictive Q-learning, even after the algorithm has converged, uses probing packets to test the status (resilience) of different routes. 
Through this process, the congested optimal path is persistently probed. 
Once the traffic level has decreased, the learning mechanism shifts back to the optimal decision path.
Special care has to be taken when the frequency of probing is selected. 
In predictive Q-learning schemes, the speed with which the packets go through a specific node - the node's recovery rate - is assumed to be unknown and estimated online.
\end{comment}

\subsection{Humans in the Loop}
There are several methods attempting to incorporate human feedback in the reinforcement learning, either to facilitate solving complex tasks,%; e.g., ones evolving in large state spaces, or ones incorporating dynamic reward indices, 
or to reshape the learning of the agent. 
%In reinforcement learning, the humans design the reward function and the learning algorithm, and after that they are no longer part of the learning scheme. 
When we add humans to the reinforcement learning loop, the human is not limited to designing the learning algorithm, but becomes part of the learning process. 
The human's role is to pass along knowledge to the agent, as we consider the environment to be complex and non stationary. 
There are many examples in the literature of how humans can intervene in the learning of the agent, like policy shaping, reward shaping, action pruning, state manipulation, etc. \cite{griffith2013internet}, \cite{ng1999internet}, \cite{wiewiora2003internet}, \cite{driessens2004internet}. 
In the context of this work, we will consider and incorporate the human into reward manipulation and action pruning of the reinforcement learning algorithm.

\subsection{Reward Shaping}
Reward shaping is a technique in which the human can change the rewards in order to influence the learning of the agent throughout the learning process. 
The human operator manipulates the environment’s rewards and, then, feeds them to the agent. 
What the agent sees is, therefore, the shaped rewards, $\overline{r}$, that the operator provides. 
The rewards the agent receives are in the form, $\overline{r} = R_A + R_H,$
%\begin{equation*}
%\overline{r} = R_A + R_H,
%\end{equation*}
where $R_A$ are the rewards that the agents receives from the environment and $R_H$ the shaped by the human operator rewards.

Algorithm $1$, given below, is the pseudocode for the proposed predictive and resilient Q-learning with human operators in the loop that manipulate the reward function.
Specifically, we define $R_A$ and $R_H$ as the rewards from the environment and the human operator respectively,  $Q^k(s_i,a_j)$ as the estimated Q-value of the state-action pair $s_i$ and $a_j$ where $i$ and $j$ are the nodes, $B^k(s_i,a_j)$ is the minimum cost incurring when in state $s_i$, action $a_j$ is taken. 
Also, $\textrm{RR}^k(s_i,a_j)$ and $U(s_i,a_j)$ are the recovery rate and the last update time, respectively, when action $a_j$ is chosen from state $s_i$. 
We use three learning parameters in the predictive and resilient Q-learning framework, $\alpha$, $\beta$, and $\gamma$. 
As in the classic Q-learning algorithm, $\alpha$ is the Q-function learning parameter, which should be equal to $1$ or the accuracy of the recovery rate might be affected. 
The recovery rate learning parameter, i.e., $\beta$, needs to obey $\beta<\gamma$, in order to regulate the decay of the recovery rate, i.e., $\gamma$, that has a direct effect on the probing frequency of a non-resilient path. 

%\begin{figure}[hbt!]
\vspace{0.3cm}

\hrule
%\vspace{0.1cm}
{\bf Algorithm $1$:} Predictive and resilient IoT Q-learning with operators in the loop performing reward manipulation
\hrule
\begin{algorithmic}%[1]
\Procedure{}{}
\small{	\State Set the $\alpha$, $\beta$, $\gamma$ parameters.
	\NoDoFor {every time window $k$} 
		\State Set environment rewards matrix $R_A^k$
		\State Update knowledge of operator %for network
		\State Human operator identifies dangerous
		\NoNumber locations, if any
		\State Initialize operator's rewards matrix $R_H^k=0$
		\NoThenIf {any dangerous or safe locations exist}
			\State Operator reshapes $R_H^k$ for dangerous 
			\NoNumber {and safe locations}
		\EndIf
		\State $R_A^k \leftarrow R_A^k + R_H^k$
		\State Initialize matrices $Q^k_\textrm{opt}$ and $B^k$ with 
		\NoNumber {sufficiently large numbers}
		\State Initialize matrices $Q^k$ and $U^k$ to zero
		\State Set the matrix $\textrm{RR}^k$ appropriately
		\NoDoFor {each epoch}
			\State Select a random initial state $s_0$
			\NoDoWhile {the goal state has not been reached}
				\State Select action $a_i$ among all possible
				\NoNumber {actions for the current state}
				\State Using this action $a_i$, consider 
				\NoNumber {going to the, next state, $s_{j}$}
				\State $\Delta Q=R^k_{A ij}+\min_{a_k}Q^k(s_j,a_k) -$ 
				\NoNumber {$Q^k(s_i,a_j)$}
				\State $Q^k(s_i,a_j) \leftarrow Q^k(s_i,a_j) + \alpha \Delta Q$
				\State $B^k(s_i,a_j) \leftarrow \min(B^k(s_i,a_j),Q^k(s_i,a_j))$
				\NoThenIf {$\Delta Q < 0$} 
					\State $\Delta \textrm{RR} \leftarrow \Delta Q /\big(\text{now}-U^k(s_i,a_j)\big)$
					\State $\textrm{RR}^k(s_i,a_j) \leftarrow \textrm{RR}^k(s_i,a_j) + \beta \Delta R$
				\Else ~{$\Delta Q>0$}
					\State $\textrm{RR}^k(s_i,a_j) \leftarrow \gamma \textrm{RR}^k(s_i,a_j)$
				\EndIf
				\State $U^k(s_i,a_j) \leftarrow \text{now}$
				\State $\Delta t = \text{now} - U^k(s_i,a_j)$
				\State $Q^k_\textrm{opt}(s_i,a_j) = \textrm{max}\big(Q^k(s_i,a_j)+$ 
				\NoNumber {$\Delta t \textrm{RR}(s_i,a_j), B^k(s_i,a_j)\big)$}
				\State Set the next state $j$ as the current 
				\NoNumber {state}
			\EndWhile
		\EndFor
		\State $y \leftarrow \textrm{arg}\min\lbrace Q^k_\textrm{opt} (s_i,a_j)\rbrace$
	\EndFor}
\EndProcedure 
\vspace{0.25cm}
\hrule
\end{algorithmic}

%\end{figure}

\section{Action Pruning}
For action pruning, the human operator observes the states that the agent comes across in the learning process and prunes actions as necessary. 
The agent is not aware that an action has been blocked by the human operator, but sees a big negative reward corresponding to that attempted action and, thus, is less likely to try it again later. 
This technique is really helpful when it comes to preventing catastrophic scenarios \cite{abel2017internet}, \cite{garcia2011internet}, \cite{garcia2015internet}, \cite{hans2008internet}. 

Algorithm $2$, given below, is the pseudocode for the proposed predictive and resilient Q-learning with human operators in the loop pruning agent's actions. 
The variables used are the same as the ones defined earlier for Algorithm $1$ on the reward shaping section.

%\begin{figure}[hbt!]

%\vspace{0.1cm}
%\begin{tabular}{l|r}
\vspace{0.3cm}
\hrule
\vspace{0.1cm}
{\bf Algorithm $2$:} Predictive and resilient IoT Q-learning with human operators in the loop performing action pruning
\hrule
\begin{algorithmic}%[1]
\Procedure{}{}
\small{	\State Set the $\alpha$, $\beta$, $\gamma$ parameters.
	\NoDoFor {every time window $k$} 
		\State Set environment rewards matrix $R_A^k$
		\State Update knowledge of human operator for network
		\State Human operator identifies dangerous, if any
		\State Initialize matrices $Q^k_\textrm{opt}$ and $B^k$ with sufficiently
		\NoNumber {large numbers}
		\State Initialize matrices $Q^k$ and $U^k$ to zero
		\State Set the matrix $\textrm{RR}^k$ appropriately
		\NoDoFor {each epoch}
			\State Select a random initial state $s_0$
			\NoDoWhile {the goal state has not been reached}
				\State Identify all possible actions
				\State Initialize human operator's rewards matrix 
				\NoNumber {$R_H^k$ to zero}
				\NoThenIf {any dangerous locations exist}
					\State Human operator reshapes $R_H^k$ for 
					\NoNumber {dangerous locations}
				\EndIf
				\State $R_A^k \leftarrow R_A^k + R_H^k$
				\State Human operator removes dangerous 
				\NoNumber {locations, if any, from possible actions}
				\State Select action $a_i$ among pruned possible 
				\NoNumber {actions for the current state}
				\State Using this action, $a_i$, consider going to the 
				\NoNumber {next state, $s_j$}
				\State $\Delta Q=R^k_{A ij}+\min_{a_k}Q^k(s_j,a_k) -$ 
				\NoNumber {$Q^k(s_i,a_j)$}
				\State $Q^k(s_i,a_j) \leftarrow Q^k(s_i,a_j) + \alpha \Delta Q$
				\State $B^k(s_i,a_j) \leftarrow \min(B^k(s_i,a_j),Q^k(s_i,a_j))$
				\NoThenIf {$\Delta Q < 0$} 
					\State $\Delta \textrm{RR} \leftarrow \Delta Q /\big(\text{now}-U^k(s_i,a_j)\big)$
					\State $\textrm{RR}^k(s_i,a_j) \leftarrow \textrm{RR}^k(s_i,a_j) + \beta \Delta R$
				\Else ~{$\Delta Q>0$}
					\State $\textrm{RR}^k(s_i,a_j) \leftarrow \gamma \textrm{RR}^k(s_i,a_j)$
				\EndIf
				\State $U^k(s_i,a_j) \leftarrow \text{now}$
				\State $\Delta t = \text{now} - U^k(s_i,a_j)$
				\State $Q^k_\textrm{opt}(s_i,a_j) = \textrm{max}\big(Q^k(s_i,a_j)+$ 
				\NoNumber {$\Delta t \textrm{RR}(s_i,a_j), B^k(s_i,a_j)\big)$}
				\State Set the next state $j$ as the current state
			\EndWhile
		\EndFor
		\State $y \leftarrow \textrm{arg}\min\lbrace Q^k_\textrm{opt} (s_i,a_j)\rbrace$
	\EndFor}
\EndProcedure 
\vspace{0.25cm}
\hrule
\end{algorithmic}

\section{Experimental Analysis}
This section demonstrates an application of the proposed predictive and resilient Q-learning algorithm for a water distribution network that incorporates human feedback for predicting the location of future leaks and forms distribution paths given a start (source) and an end (destination) node. 
The proposed path avoids, if any, dangerous locations according to the human, and involves as few as possible locations with leaks. 
The demonstration relies on real data of the leaks that occurred over the last five years in the Arlington County, Virginia. 
We want to make sure that no matter what happens (possible attacks/sabotage or leaks on the pipelines of the water network of the County), water from the assumed source which is by the bank of river Potomac, namely neighborhood $1$, reaches the Ronald Reagan Washington Regional Airport, which is the destination corresponding to neighborhood $119$.

For the purposes of this experiment, we are considering the scenario in which the Arlington’s Pediatric Centers have admitted multiple children with stomachaches, and in some cases severe gastroenteritis. 
The number of admitted children is alarming, and since the children share no common background %(e.g. attending the same school or living in the same neighborhood) 
the Pediatric Centers have shared information about the incidents with the water distribution authorities of the County. 

A warning is given out to the local water utilities to investigate potential contamination as the cause of the gastroenteritis outbreak. 
The patients' demographic data are processed and utilities employees are sent to the suspected locations to collect and test water samples. 
However, the preliminary analysis of the water samples takes several hours and the full analysis at least a day. 
While waiting for the test results, the water distribution authorities want to minimize the amount of water passing through the risky locations to prevent the spread of a possible water contamination. 

The scenario described above resembles several past cases of major water contamination. 
In $1990$ Cabool, Missouri, experienced a large outbreak of Escherichia coli, same as the one in $2000$ in the town of Walkerton, Ontario, Canada. 
In $1993$, more than $1.61$ million residents in Milwaukee, Wisconsin, became ill because of the Cryptosporidium protozoan that that had passed through the filters in the water system in the area \cite{clark2004internet}. 
Given these examples and many other unintentional water contamination incidents in history, the water distribution authorities do not want to take risks, but shutting down the water before strong evidence of contamination is available is not well-tolerated by businesses and residents either.

When the results of the tests are available, the severity of the situation and the possible contamination, as well as the necessary actions, such as issuing a warning, temporarily isolating the dangerous locations from the network, can be determined.

In such scenarios, human operators have to interrupt the learning and provide information in some form to the agent about the new data or state of the network that are not accessible by the agent.
This work precisely showcases the interaction between human operators and agents in order to exercise this preventative measures.

\subsection{Dataset}
As previously mentioned, the dataset in this study contains $1816$ instances of leaks in the water network of the Arlington County, over the last five years. 
Each instance involves information about the location of the leak, the time period between the identification of the leak and its repair, as well as the occurred cost.

For the purposes of this work, the geographic area of Arlington County is divided into $119$ neighborhoods, as shown in Figure~$4$ and Figure~$5$ %and Figure~$6$. 
%We are assuming that neighborhood $1$ corresponds to the source of the water network of the County, since it is by the bank of river Potomac. 
%In addition, we assume that the Ronald Reagan Washington Regional Airport, which corresponds to neighborhood $119$, is the destination of the water network. 
%These two neighborhoods ($1$ and $119$) are considered to have no leaks and to be impermeable to possible attacks. 

As in our previous work \cite{grammatopoulou2018internet}, the data of the location instances are classified into $117$ sets depending on their location ID. 
%Each of the $117$ sets corresponds to a geographic area of the Arlington County.
Because some neighborhoods are more vulnerable to leaks and attacks than others, the total number of leaks appearing in each neighborhood is different. 
%The rewards of the environment that are assigned to the edges of the network connecting the neighborhoods are determined by using all the provided information except the location of the leaks. 

When two neighborhoods share a border, we consider them to be connected with a direct pipeline. 
The states and actions of the proposed predictive and resilient Q-learning algorithm are defined using the former notion. 
The state-action pair $(s_i,\alpha_j)$ denotes that while being on neighborhood $i$, we choose to move to neighborhood $j$, which is directly connected to neighborhood $i$. 
Therefore, $Q(s_i,\alpha_j)$ is the cost of channeling the water from neighborhood $i$ to neighborhood $j$. 
The problem becomes more computationally complex as the number of neighborhoods sharing a border with other neighborhoods gets larger, since the same happens with the action space grows with respects to to that state. 

\subsection{Training Details} 
The proposed framework uses time windows, with each time window handling M = $30$ data instances. 
First the environment’s reward matrix for the time window is computed, accounting not only for the current number of leaks, the time to repair them, and the cost produced by them, but also the historical values in an exponentially decreasing way for each neighborhood . 
Subsequently, the data history concerning the network is transmitted to the human operator.
If the operator has knowledge of changes in the status of the network that render manipulation of the learning necessary, he interacts with the agent to pass on the new information. 
To assess the efficacy of the algorithm in highly unstructured environments, the changes in the status of the network concerning dangerous and safe locations are randomly generated.
Figure~$1$ shows the number of neighborhoods having leaks, together with those indicated as dangerous or safe by the human operator in different time windows. 
Following that, the agent receives information both from the environment regarding the location of leaks (faults), as well as status of the network (e.g., suspected contamination) from the human operator, and the training session starts, in which the system is trained for $100$ epochs.

\begin{figure*}%[hbt!]
\centering
\includegraphics[width=2\columnwidth,height=8cm]{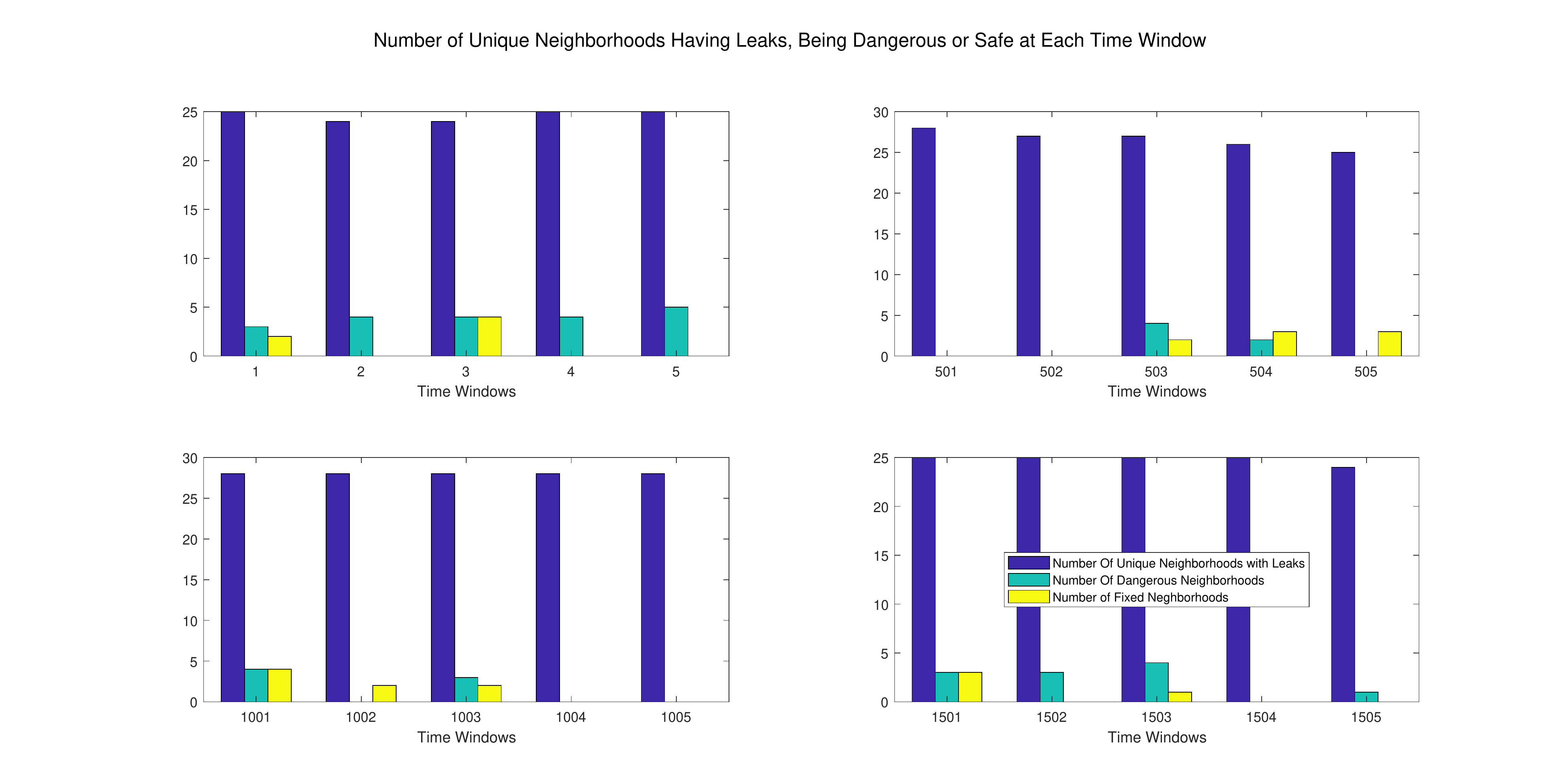}
\caption{Number of neighborhoods having leaks, and being indicated as danger or safe at different time windows..}~\label{fig:figure1}
\end{figure*}

Once the training for the time window is completed, the location of the possible future leaks is predicted, and a path is proposed for connecting the source with the destination, involving as few as possible neighborhoods with leaks. 
The rationale behind the selection of the neighborhoods for the proposed path is that we want to reach our destination incurring the minimum cost. 
To achieve that, the selection of the nodes is based on the $Q_\textrm{opt}$ matrix of the time window, which contains the cost for transitioning from one neighborhood to another. 
The operator-indicated dangerous locations are avoided at all costs in the formation of the proposed optimal path. 
For the neighborhoods having leaks and not being part of the proposed path, the agent suggests isolation from the water network to the human operator for initiating repair of the pipelines and reducing the costs. 
As for the neighborhoods indicated as dangerous, their isolation, until the results of the water tests become available, is also proposed. 
By isolating the parts of the network that might have been compromised, the authorities are given more time to run the necessary tests for evaluating the severity of the situation, examine and the need to issue public warning, and diagnose whether the gastroenteritis outbreak in the children was originated from bacteria in the water.

With a significant amount of training data, the absolute difference between the values of two consecutive $Q_\textrm{opt}$ matrices should approach zero. 
As illustrated in Figure~$2$, the absolute value of the difference of the $Q_\textrm{opt}$ matrices between two consecutive time windows in our system converges to zero early on the simulation, specifically at the 50th time window. 
The fluctuations observed in Figure~$2$ are caused by two things: (i) the fact that leaks in time window $(k+1)$ are appearing in different neighborhoods than in time window $k$, and (ii) the randomness in frequency and number of the dangerous and safe neighborhoods indicated by the human operator. 
Since our system is not trained on these specific scenarios, the predictions of the locations of the future leaks differs from the actual ones. 

\begin{figure}%[hbt!]
\centering
\includegraphics[width=0.9\columnwidth]{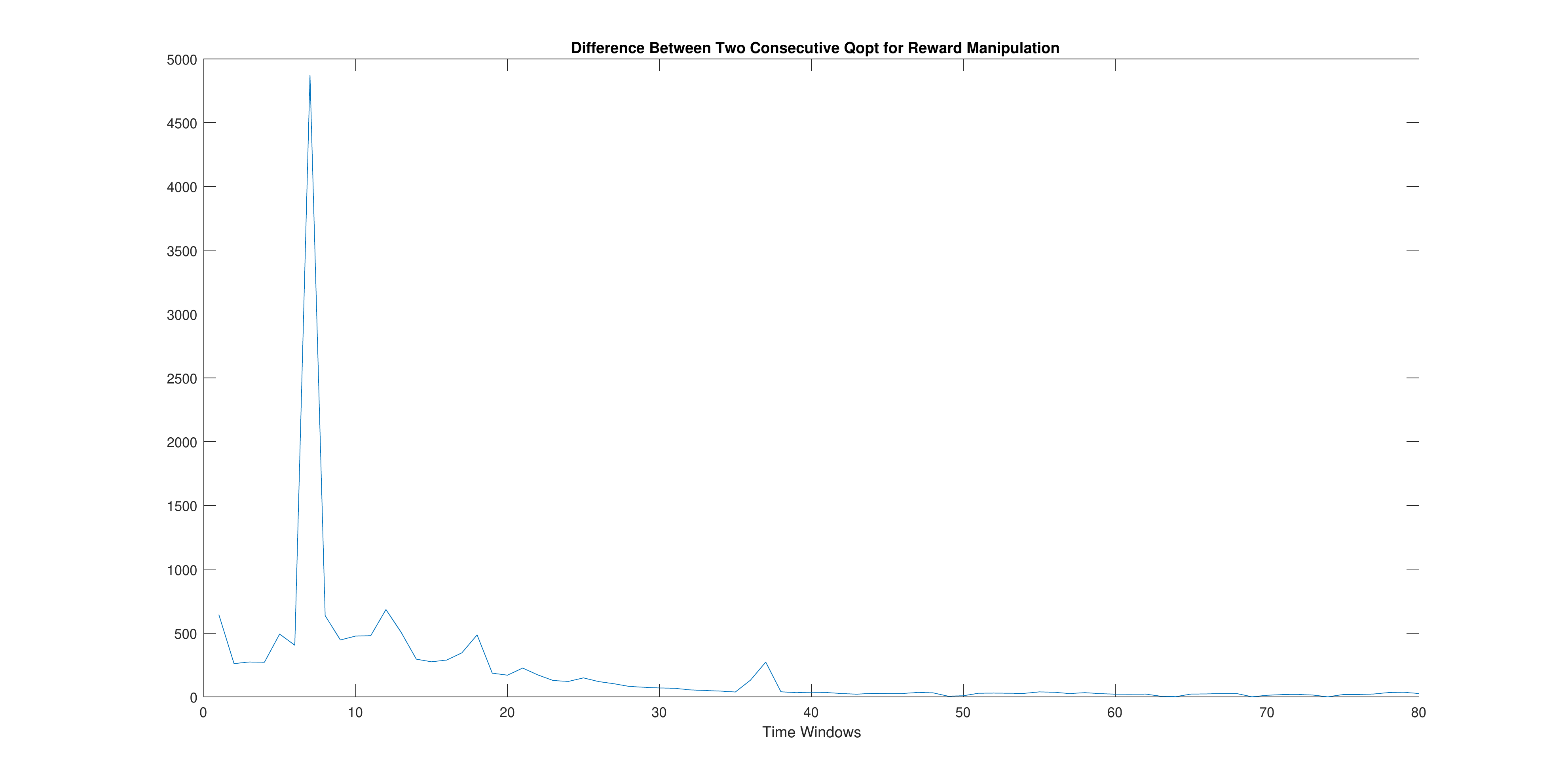}
\caption{Difference between two consecutive $\mathbf{Q_\textrm{opt}}$ matrices over time.}~\label{fig:figure2}
\end{figure}

Figure~$3$ illustrates how the absolute difference between the values of two consecutive $Q_\textrm{opt}$ matrices for the simple Q-learning, Q-learning with reward manipulation and action pruning for the first $70$ time windows. 
The intervention of the human with the learning of the agent has an impact on the  $Q_\textrm{opt}$ matrices. 
The differences between the plot of the reward manipulation and the action pruning are due to the fact that the human identifies both dangerous and safe locations in the former, but only dangerous ones in the latter. 
Although the fluctuations in the absolute difference between the values of two consecutive $Q_\textrm{opt}$ matrices in the first time windows are big, Figure~$2$ indicates that they approach zero subsequently.

\begin{figure}%[hbt!]
  \centering
  \includegraphics[width=0.9\columnwidth]{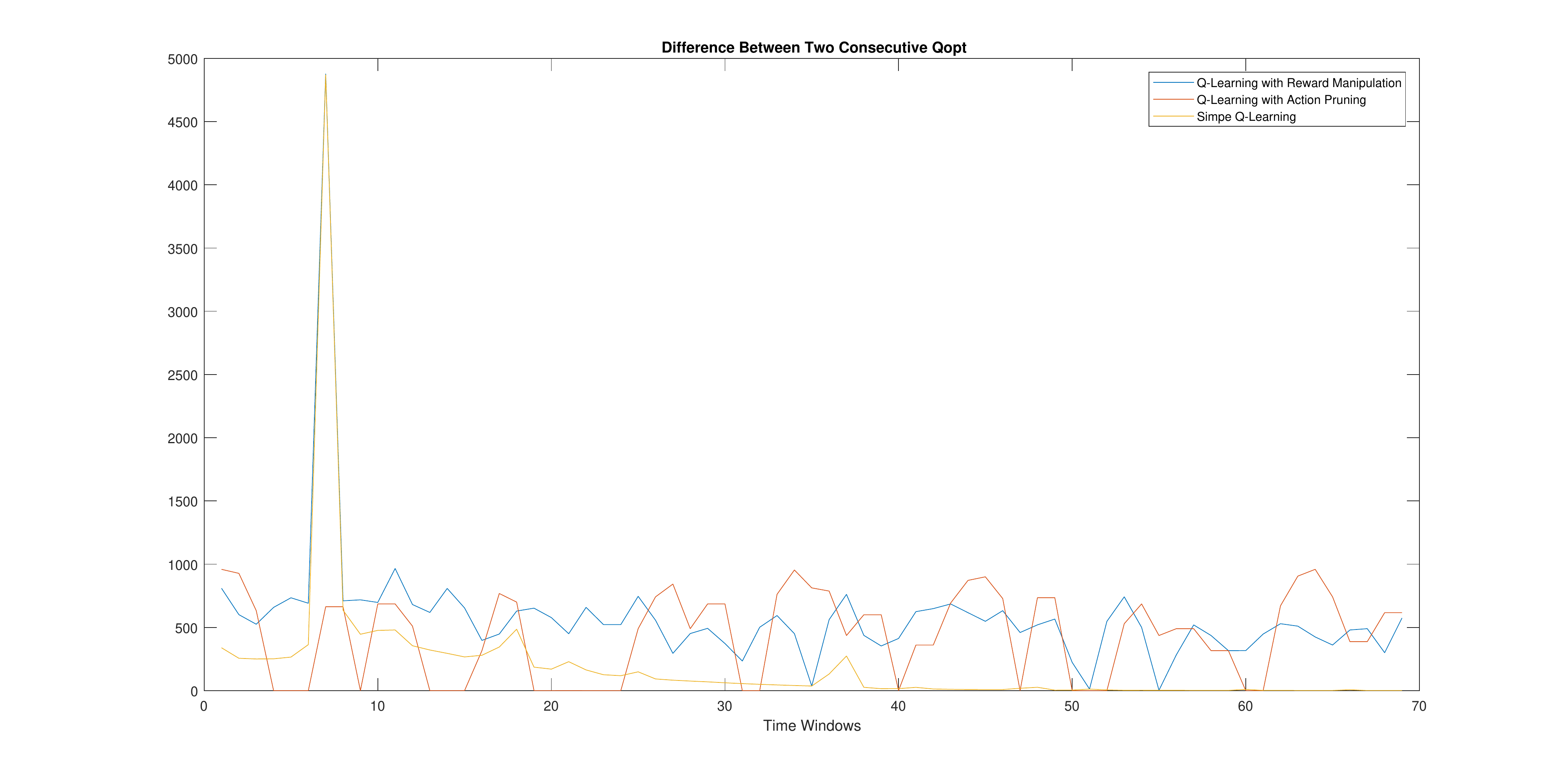}
  \caption{Difference between two consecutive $\mathbf{Q_\textrm{opt}}$ matrices for simple Q-learning, and Q-learning with reward manipulation and action pruning over time.}~\label{fig:figure3}
\end{figure}

\subsection{Results}
The predictions of the system, as expected, improve with more training windows.
In order to have a better understanding of the system, and how it evolves over time, we examine the results of two different time windows, randomly chosen, the $512^{th}$ %and $1726^{th}$ 
for reward shaping and the $236^{th}$ for action pruning. 

Figure~$4$ shows the nodes having leaks (purple colored), the nodes that the human operator indicated as dangerous (red colored), and the proposed optimal path (marked with blue stars) at the end of the $236^{th}$ time window for the method of action pruning. 
In action pruning, the operator can only identify new neighborhoods as dangerous. 
%The proposed path started from the source (neighborhood $1$), and passing through neighborhoods $4$, $10$, $17$, $29$, $38$, $59$, $80$, $89$, $95$, $100$, $105$, $107$, $111$ and $118$, reached the destination (neighborhood $119$). 
%One might be confused as to why neighborhood $59$, which had some leaks, was part of the proposed optimal path. 
%The reason is that the path is sometimes more cost effective to employ a neighborhood with leaks than not, especially when we are aiming at controlling the water flow in the events of a potential contamination.
%We can infer this also from Figure~$5$, which shows the cost of the optimal paths that employed neighborhoods with leaks in several time windows.

\begin{figure}%[hbt!]
\centering
\includegraphics[width=\columnwidth, height=0.8\columnwidth]{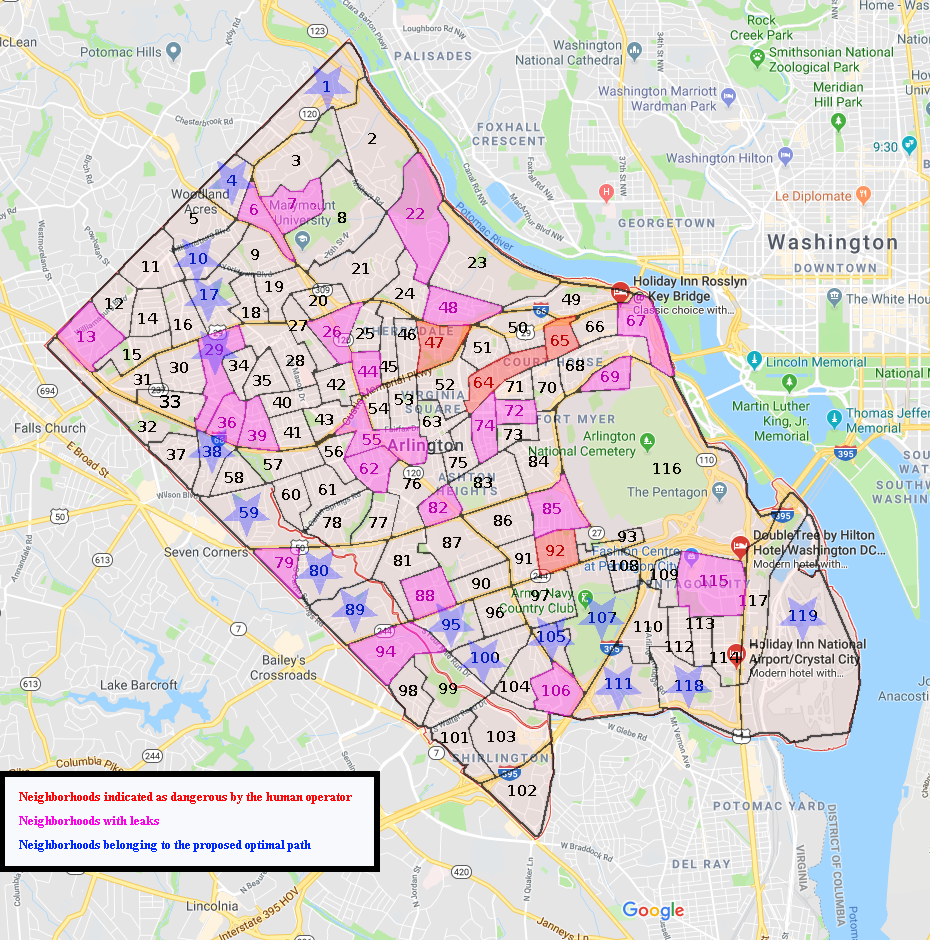}
\caption{Optimal proposed path and labeled neighborhoods after the $\bold{236^{th}}$ time window for action pruning.}~\label{fig:figure4}
\end{figure}

\begin{figure}%[hbt!]
\centering
\includegraphics[width=0.9\columnwidth]{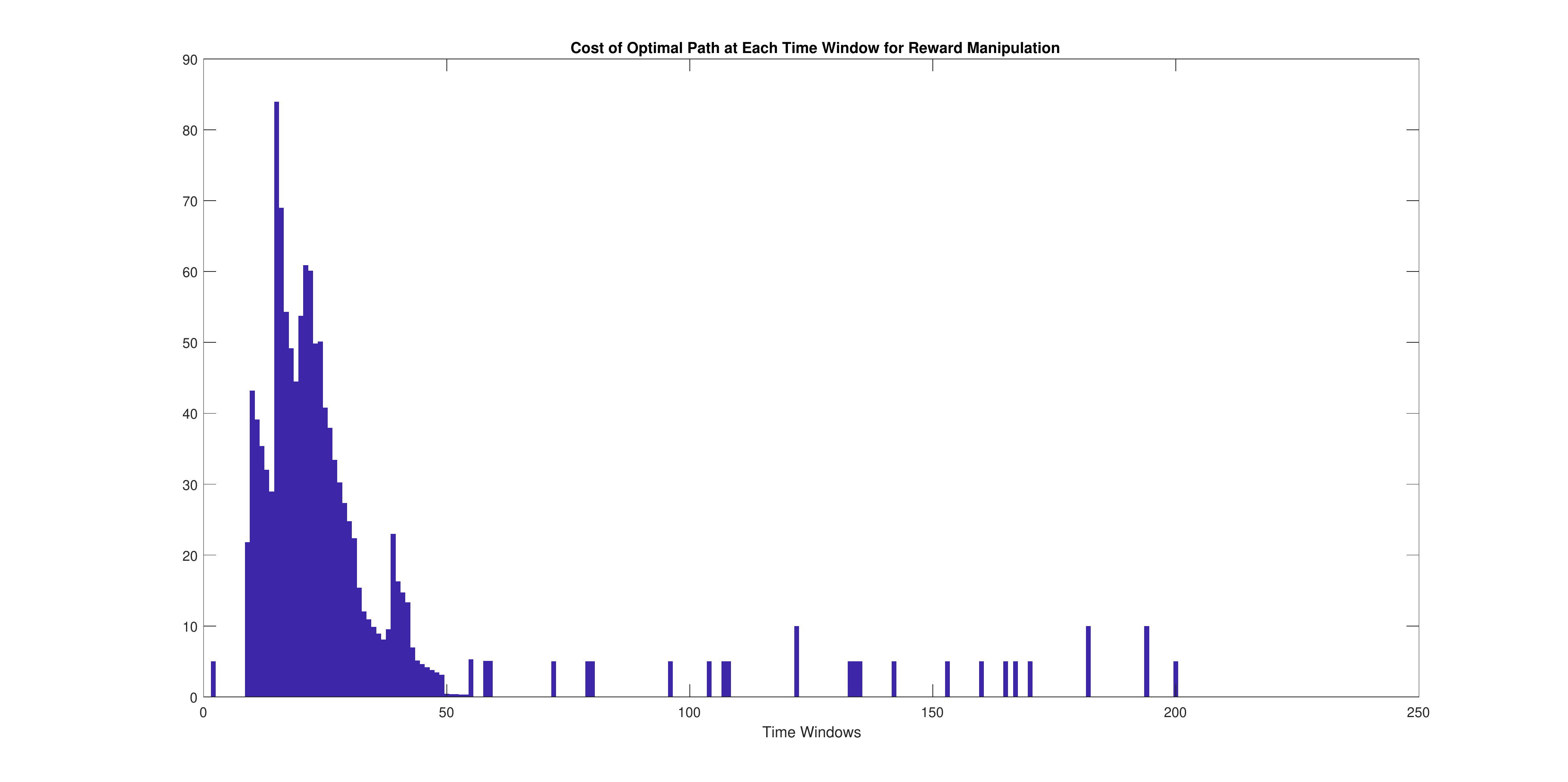}
\caption{Cost of the optimal path due to the appearance of leaks.}~\label{fig:figure5}
\end{figure}

Similarly, Figure~$6$ shows the nodes having leaks (purple colored), the nodes that the human operator indicated as dangerous or safe (red and green colored respectively), and the proposed optimal path (marked with blue stars) at the end of the $512^{th}$ time window for the method of reward shaping. 
Unlike action pruning, reward shaping permits the human operator to identify new locations as dangerous and to re-evaluate the status of previously classified ones. 
%The proposed path, following the logic in the analysis before, employed exactly the same nodes as the previous example. 
%However, if we pay closer attention we notice that the status of the neighborhoods differed. 
%The reason why the algorithm found the same optimal path in two different time windows, might have been purely by chance. However, the results might have been driven by more important reasons: that these neighborhoods were less vulnerable to leaks as others and therefore their $Q_\textrm{opt}$ cost values were smaller than others’ as illustrated in Figure~$2$; or that these neighborhoods were not associated with any potential water contamination. 
%Note that neighborhood $29$ was among those that the operator indicated as safe. 
%This means that there had been concerns about potential infection of the water at the specific neighborhood in the previous time window, but at the current time window the possibility of contamination in the neighborhood, given the test results, was ruled out. 

\begin{figure}%[hbt!]
\centering
\includegraphics[width=0.9\columnwidth,height=0.8\columnwidth]{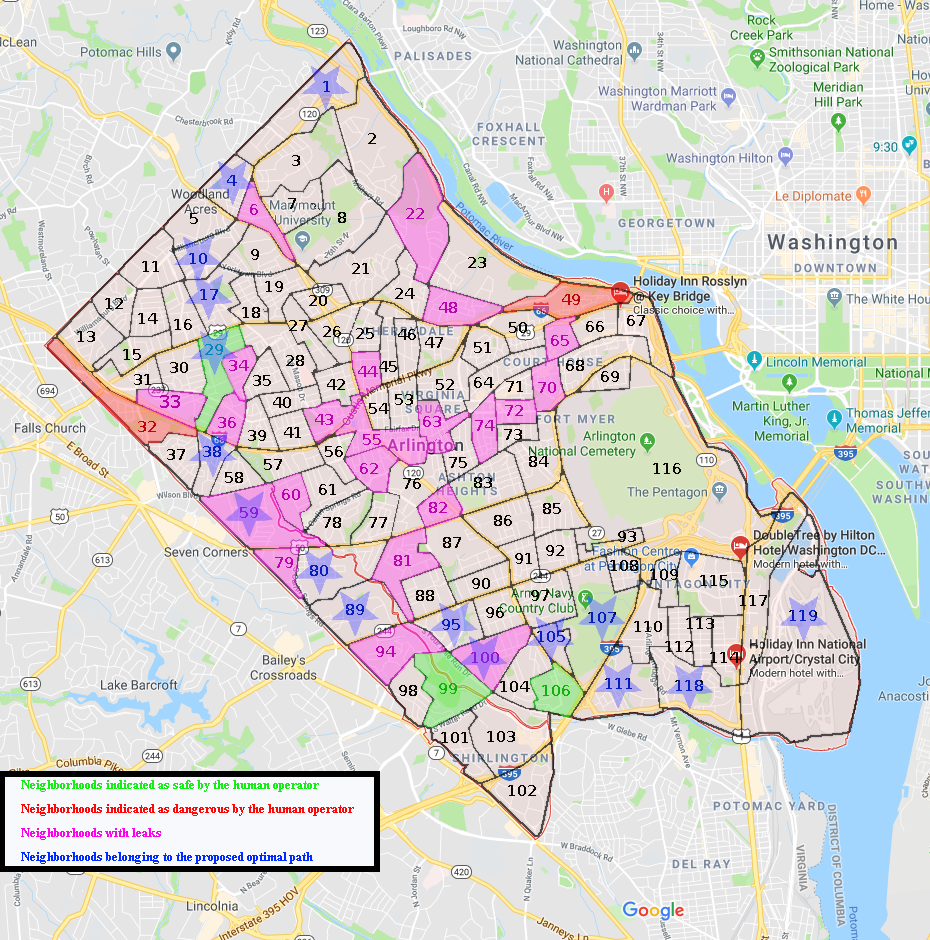}
\caption{Optimal proposed path and labeled neighborhoods after the $\bold{512^{th}}$ time window for reward manipulation.}~\label{fig:figure6}
\end{figure}

% Similarly, Figure~$7$ shows the nodes having leaks, the nodes that the human operator indicated as dangerous or safe, and the proposed optimal path at the end of the $1726^{th}$ time window for the method of reward manipulation. 
% The proposed path started from the source (neighborhood $1$), and passing through neighborhoods $4$, $10$, $17$, $18$, $19$, $27$, $42$, $44$, $54$, $55$, $76$, $87$, $90$, $97$, $107$, $111$ and $118$, reached the destination (neighborhood $119$). 
% In this time window the location of the neighborhoods having leaks changed significantly compared to the previously mentioned time windows and so did the proposed optimal path. 
% In order for the dangerous neighborhoods to be isolated from the water network, we see that the optimal path included two locations with leaks, namely $27$ and $107$.

% \begin{figure}%[hbt!]
% \centering
% \includegraphics[width=\columnwidth, height=0.8\columnwidth]{1726RM}
% \caption{Optimal proposed path and labeled neighborhoods after the 
% $\bold{1726^{th}}$ time window for reward manipulation.}~\label{fig:figure7}
% \end{figure}
 
Looking at Figures~$4$ and $6$%, and $7$ together
, one can observe that some neighborhoods with leaks that are being isolated from the rest of the network are common among the two time windows. 
A neighborhood may be suggested for isolation in multiple time windows for three reasons: (a) the leak in the pipelines has not been fixed yet, (b) vulnerabilities and leaks keep appearing in the neighborhood, or (c) the human marks the location as dangerous while the results of the water tests are not yet ready. 
%In the first and third case, the neighborhood would be among the isolated ones in two consecutive time windows, but this would not necessarily be true in the second case.

\section{Discussion} %added by nathan that need further discussion.
This study presents reinforcement learning algorithms that support action pruning and reward manipulation by the human in real time to adapt the agent's actions on an IoT network. 
By definition, an IoT network cannot be fully prepared for unanticipated events, which often involve information that cannot be processed by the agent; thus, the human becomes an essential component in the system to interpret these information and adapt the actions of the agent. 
%Through our case study of the water supply network in Arlington County, VA, 
We have shown that human in the loop action pruning and reward manipulation are both feasible and important for adapting automatic control of an IoT network to unanticipated situations. 
%Specifically, our simulation results indicate new water distribution paths for avoiding potentially contaminated sites given simulated human interventions (i.e., action pruning or reward manipulation). 
%At the same time, the agent is able to provide paths that account for all the historical information on leaks and repair. %expand on result discussion but I am a bit limited in interpreting the results

In our case study, reward manipulation and action pruning yield very similar optimal paths and thus are equally effective in terms of physically mitigating the spread of contamination. However, action pruning is less strenuous for the human operator and thus is preferred. 

\section{Conclusion}
We present two algorithms -- Q-learning with reward manipulation and action pruning -- that allow interaction between the human operators and the agents in large-scale IoT networks in order to  recommend resilient policies for scheduling problems, as well as to optimally implement containment and recovery actions as necessary. 
We model the IoT as a graph, and we integrate data from recorded node failures in real-time by updating the cost matrix of the graph and, thus, dynamically shifting the optimal path choice. 
In addition, we introduce two ways that a human operator can interact with the agent by manipulating its learning, based on last minute changes in the environment and the status of the network only perceivable by the human (i.e., information from sources/sensors that is not available or integrated into the network). 
We %use real anonymized data provided by the Arlington County, Virginia, USA, to
highlight the effects of malicious, random water network failures, and simulate the possibility of malicious or non-random contamination of the water in certain nodes to showcase the interaction between human and machine.
Future work will focus on applying the developed framework to different scenarios, exploring more ways for the human operators to interact with the agent.

% BALANCE COLUMNS
\balance{}

\bibliographystyle{IEEEtran}

\bibliography{IOT_2019_references}

\end{document}